\documentclass[letterpaper, 10 pt, conference]{ieeeconf}  

\IEEEoverridecommandlockouts                              

\usepackage{cite}
\usepackage{amsmath,amssymb,amsfonts}
\usepackage{graphicx}
\usepackage[table]{xcolor} 
\usepackage{multirow} 
\usepackage{makecell} 
\usepackage{float}
\usepackage{pax}
\usepackage{hyperref}
\usepackage{url} 
\usepackage{lipsum}
\usepackage{siunitx}

\usepackage{color,soul}

\newcommand{\mb}[1]{\mathbf{#1}}

\title{\LARGE \bf One Net to Rule Them All: Domain Randomization in Quadcopter Racing Across Different Platforms}

\author{Robin Ferede, Till Blaha, Erin Lucassen, Christophe De Wagter, Guido C.H.E. de Croon
\thanks{$^{1}$The authors are with the Micro Air Vehicle Lab of the Faculty
of Aerospace Engineering, Delft University of Technology, 2629 HS
Delft, The Netherlands {\tt \scriptsize robinferede@tudelft.nl, t.m.blaha@tudelft.nl, e.lucassen@student.tudelft.nl, c.deWagter@tudelft.nl, g.c.h.e.deCroon@tudelft.nl}}%
}

\begin{document}

\maketitle
\thispagestyle{empty}
\pagestyle{empty}

\begin{abstract}
In high-speed quadcopter racing, finding a single controller that works well across different platforms remains challenging. This work presents the first neural network controller for drone racing that generalizes across physically distinct quadcopters. We demonstrate that a single network, trained with domain randomization, can robustly control various types of quadcopters. The network relies solely on the current state to directly compute motor commands. The effectiveness of this generalized controller is validated through real-world tests on two substantially different crafts (3-inch and 5-inch race quadcopters). We further compare the performance of this generalized controller with controllers specifically trained for the 3-inch and 5-inch drone, using their identified model parameters with varying levels of domain randomization (0\%, 10\%, 20\%, 30\%). While the generalized controller shows slightly slower speeds compared to the fine-tuned models, it excels in adaptability across different platforms. Our results show that no randomization fails sim-to-real transfer while increasing randomization improves robustness but reduces speed. Despite this trade-off, our findings highlight the potential of domain randomization for generalizing controllers, paving the way for universal AI controllers that can adapt to any platform.
\end{abstract}

\section{INTRODUCTION}
The drone market is experiencing rapid growth \cite{Hassanalian2017ClassificationsAA}, yet most applications still rely on human pilots. Given that battery life limits flight times, enhancing the speed and autonomy of drones offers significant advantages \cite{hanover_autonomous_2024}. 
As a result, research into autonomous drone racing is a growing field that contributes to advancing technology and enhancing drone capabilities.

One of the significant challenges in drone racing is the development of controllers that generalize well across different quadcopter platforms. While human pilots can easily adapt to various drones with minimal adjustments, current racing AI systems often overfit to their specific platforms, restricting their generalization capabilities. This limitation hampers the transfer of current drone racing technology to other real-world applications \cite{hanover_autonomous_2024}.

Recent advancements in quadcopter control have predominantly focused on Reinforcement Learning (RL) techniques \cite{Autonomous_Drone_Racing_with_Deep_Reinforcement_Learning, penicka2022learning, OCvsRL, Kaufmann2023, eschmann_learning_2023}. These methods have demonstrated impressive results in drone racing, with neural controllers surpassing human champions \cite{Kaufmann2023}. However, these achievements rely on precise modeling of the drone platforms and struggle to generalize across different drones.

\begin{figure}
    \centering
    \includegraphics[width=\linewidth]{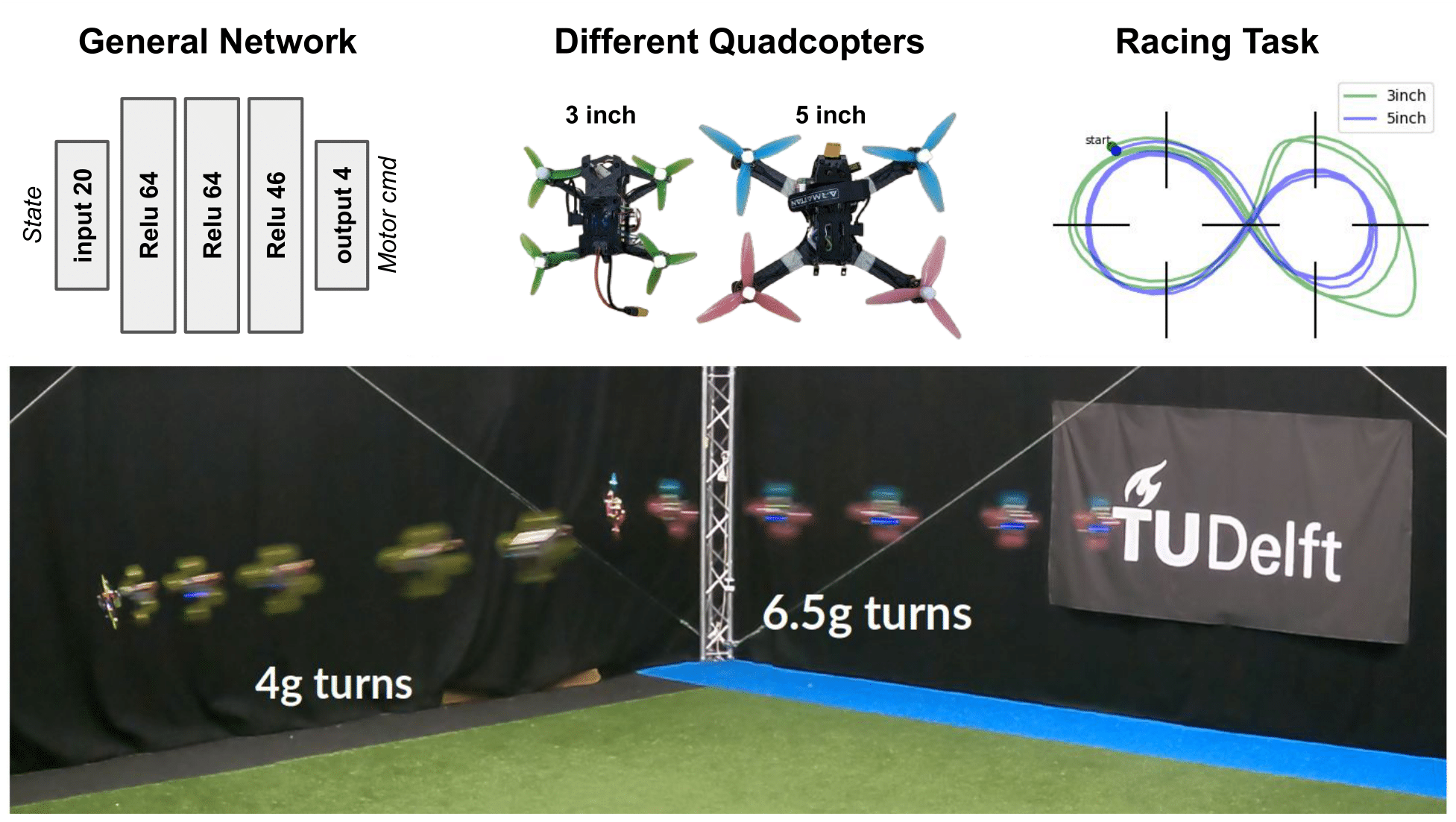}
    \caption{We train a single neural network to control physically distinct drones. The time-lapse image demonstrates the successful sim-to-real transfer of the reinforcement-learned network, enabling it to perform drone racing with both 3-inch and 5-inch quadcopters.}
    \label{fig:fig1}
    \vspace{-4mm} 
\end{figure}
Domain randomization (DR) has emerged as a prominent method for transferring RL policies across different environments. This technique involves training policies in simulated environments with varied parameters to improve their robustness when deployed in real-world scenarios \cite{josifovski2022analysis}. For instance, DR has facilitated single-shot sim-to-real transfer for robot manipulation \cite{andrychowicz2020learning}, visual navigation using synthetic data \cite{loquercio2019deep}, and object detection \cite{tobin2017domain}. Additionally, in end-to-end quadcopter control, DR, combined with real-time adaptation, has been critical for bridging the reality gap \cite{robinSl, Seb, robinRL}. However, although DR enhances robustness, it comes with trade-offs, including longer training times and diminished performance optimality \cite{tiboni2023domain, josifovski2022analysis}.

Current research on generalized quadcopter control has only focused on low-level tasks in near-hover or recovery-to-hover conditions, with little exploration into racing scenarios that demand both guidance and control at significantly higher speeds. Recent studies have explored various approaches: one developed a network for stabilizing low-level control (position control and trajectory tracking) that generalizes to various quadcopters using domain randomization \cite{molchanov2019sim}. Another method \cite{blaha_control_2024} introduced a single-shot real-time technique for controlling an unknown quadcopter through iterative determination of the actuator effectiveness parameters of an INDI controller, expanding on the work of \cite{smeur_adaptive_2016}. Additionally, recent work trained a neural network to estimate a latent representation of quadcopter parameters, conditioning another neural network controller for near-hover flight, enabling online adaptation to varying hardware and disturbances during real-world deployment \cite{zhang_learning_2023}.

In this work, we extend these results to drone racing by using domain randomization to develop a neural network controller for high-speed racing that generalizes across different quadcopters. Our controller maps state information directly to motor commands, enabling end-to-end guidance and control. Our key contributions include:
\begin{enumerate}
    \item Introducing, to the best of our knowledge, the first neural network controller capable of operating across different drones for high-speed racing
    \item Investigation into the effects of varying degrees of domain randomization on sim-to-real transferability and flight speed performance.
\end{enumerate}
This paper is organized as follows:  Sec. \ref{section:methods} presents the drone model and control formulation; Sec. \ref{section:experiment} describes the experimental setup; Sec. \ref{section:results} analyzes the performance of the general control network and examines the effects of varying degrees of domain randomization; and Sec. \ref{sec:conclusion} provides the conclusion and outlines directions for future work.

\section{METHODOLOGY}\label{section:methods}
\subsection{Quadcopter model}
To train our neural policies we simulate our quadcopter dynamics with a parametric model similar to the one outlined in \cite{robinRL}. The quadcopter's state and control are given by:
\begin{align*}
    \mb{x} = [\mb{p}, \mb{v}, \boldsymbol \lambda, \mb{\Omega}, \mb{\boldsymbol \omega}]^T \quad \mb{u} = [u_1, u_2, u_3, u_4]^T
\end{align*}
Here $\mb{p}$ is the position of the drone, $\mb{v}$ is the velocity, $\boldsymbol \lambda$ are the Euler angles, $\bold \Omega$ are the body rates and $\boldsymbol \omega$ are the propeller speed in rad/s. $\mb{u} \in [0,1]^4$ represents the normalized commands sent to the motor controllers. The equations of motion are expressed as follows:
\begin{align}
    \dot{\mb{p}} &= \mb{v} &
    \dot{\mb{v}} &= g \mb{e_3} + R(\mb{\lambda}) \mb{F}\\
    \dot{\mb{\lambda}} &= Q(\mb{\lambda}) \mb{\Omega} &
    \dot{\mb{\Omega}} &= \mb{M} \nonumber \\
    \dot{\omega}_i  &= (\omega_{ci} - \omega)/\tau &
\end{align}
Where $R$ is the rotation matrix and 
$Q$ transforms body rates to Euler angle derivatives. The steady state motor response is modeled as: $\omega_{ci}=(\omega_{\text{max}}-\omega_{\text{min}})\sqrt{k_l u_i^2 + (1-k_l)u_i} + \omega_{\text{min}}$. The specific force $F$ and the moment $M$ \footnote{$M$ represents angular acceleration. The moment of inertia is inderectly estimated throught the $k_{\square}$ and gyroscopic effects are ignored.} are modeled as:
\begin{align*}
    \mb{F} = \Big[
        - \sum_{i=1}^4  k_x v^B_x \omega_i,
        - \sum_{i=1}^4  k_y v^B_y \omega_i,
        - \sum_{i=1}^4  k_\omega \omega_i^2 \Big]^T
\end{align*}
\begin{align*}
    \mb{M} = \begin{bmatrix}
        -k_{p1} \omega_1^2 -k_{p2} \omega_2^2 + k_{p3} \omega_3^2 + k_{p4} \omega_4^2\\
        -k_{q1} \omega_1^2 + k_{q2} \omega_2^2 - k_{q3} \omega_3^2 + k_{q4} \omega_4^2\\
        -k_{r1} \omega_1 + k_{r2} \omega_2 + k_{r3} \omega_3 - k_{r4} \omega_4 + \\
        ... -k_{r5} \dot{\omega}_1 + k_{r6} \dot{\omega}_2 + k_{r7} \dot{\omega}_3 - k_{r8} \dot{\omega}_4
    \end{bmatrix}
\end{align*}
The parameters $k_{\square}, \omega_{\text{min}}, \omega_{\text{max}}$ are estimated from manual flight data for both the 3-inch and 5-inch quadcopters using linear regression. We normalize the parameters based on the maximum angular velocity \( \omega_{\text{max}} \). The normalized parameters are:
\[
\hat{k}_{\omega},\hat{k}_{pi},\hat{k}_{qi} = (k_{\omega},k_{pi},k_{qi}) \times \omega_{\text{max}}^2
\]
\[
\hat{k}_{xi},\hat{k}_{yi},\hat{k}_{ri} = (k_{xi},k_{yi},k_{ri}) \times \omega_{\text{max}}\\
\]
This scaling accounts for variations in \( \omega_{\text{max}} \) and ensures consistency across the model. The estimated values can be found in Tab. \ref{tab:param}. These parameters highlight the distinct control characteristics of each platform. For instance, the 3-inch platform has approximately half the thrust-to-weight ratio (proportional to $\hat{k}_{\omega}$), motors that spin 50\% faster, and less than half the pitch effectiveness ($\hat{k}_{pi}$) compared to the 5-inch platform.
\begin{table}
\centering
\scriptsize
\caption{Parameters identified for 3inch and 5inch drone}
\begin{tabular}{c|c|c||c|c|c}
\hline
\hline
param & 3 inch & 5 inch & param & 3 inch & 5 inch \\
\hline
$\hat{k}_{\omega} $ & 14.3 & 27.1 & $\omega_{\text{min}}$ (rad/s) & 305.4 & 238.49 \\
$\hat{k}_{x} $ & 0.16 & 0.16 & $\omega_{\text{max}}$ (rad/s) & 4887.57 & 3295.5 \\
$\hat{k}_{y} $ & 0.18 & 0.24 & $k_l$ & 0.84 & 0.95 \\
    & & & $\tau$ (sec) & 0.04 & 0.04 \\
\hline
$\hat{k}_{p1}$ & 615.0 & 711.0 & $\hat{k}_{r1}$ & 47.1 & 35.2 \\
$\hat{k}_{p2}$ & 598.0 & 718.0 & $\hat{k}_{r2}$ & 47.1 & 35.2 \\
$\hat{k}_{p3}$ & 650.0 & 691.0 & $\hat{k}_{r3}$ & 47.1 & 35.2 \\
$\hat{k}_{p4}$ & 479.0 & 724.0 & $\hat{k}_{r4}$ & 47.1 & 35.2 \\
\hline
$\hat{k}_{q1}$ & 217.0 & 573.0 & $\hat{k}_{r5}$ & 5.57 & 6.49 \\
$\hat{k}_{q2}$ & 238.0 & 637.0 & $\hat{k}_{r6}$ & 5.57 & 6.49 \\
$\hat{k}_{q3}$ & 280.0 & 548.0 & $\hat{k}_{r7}$ & 5.57 & 6.49 \\
$\hat{k}_{q4}$ & 196.0 & 640.0 & $\hat{k}_{r8}$ & 5.57 & 6.49 \\
\hline
\end{tabular}
\label{tab:param}
\end{table}

\subsection{RL problem definition} \label{sec:RLproblem}
The racing setup consists of seven square gates, each measuring 1.5x1.5 meters, arranged in a figure-eight track, as shown in Fig.~\ref{fig:fig1}. The drone starts 1 meter in front of a randomly selected gate, in a similar fashion as \cite{Autonomous_Drone_Racing_with_Deep_Reinforcement_Learning}. The other initial state variables are defined as follows: $\mb{v} \in [-0.5, 0.5]^3$, $\phi, \theta \in [-\frac{\pi}{9}, \frac{\pi}{9}]$, $\psi \in [-\pi, \pi]$, $\mb{\Omega} \in [-0.1, 0.1]^3$, and $\mb{\omega} \in [\omega_{\text{min}}, \omega_{\text{max}}]$. Similar to \cite{OCvsRL} the reward function to be optimised consists of a progress reward, a rate penalty and a collision penalty:
\begin{align*}
    r_k = \begin{cases}
        -10, &\text{if collided} \\
       ||\mb{p}_{k-1} - \mb{p}_{g_k}||-||\mb{p}_{k} - \mb{p}_{g_k}|| - c||\mb{\Omega}|| &\text{otherwise}
    \end{cases}
\end{align*}
Here, \(\mb{p}_{g_k}\) denotes the position of the center of the current target gate, while \(\mb{p}_{k}\), \(\mb{p}_{k-1}\) are the drone's current and previous positions. The rate penalty coefficient is set to $c=0.001$. A collision is registered when the drone either makes contact with the ground or when it flies outside of a 10x10x7m bounding box. When a collision is registered, the episode ends. Additionally when a gate is missed (i.e. drone passes gate plane outside of 1.5x1.5m area, the episode also ends. 

This reward function, together with the quadcopter model from the previous section, is used to create a gym environment for training with the PPO algorithm \cite{ppo}, utilizing the Python library Stable-Baselines3 \cite{stable-baselines3}. Our implementation simulates 100 drones in parallel, uses a discount factor of $\gamma=0.999$ and a maximum episode length of 1200 time steps (12 seconds). The training process concludes after a total of 100 million time steps. The code for our implementation is openly available on GitHub\footnote{\url{https://github.com/tudelft/optimal_quad_control_RL/tree/icra2025}}.

\subsection{Randomization}
\subsubsection{General policy} \label{sec:general}
The randomization outlined in Tab. \ref{tab:randomization_parameters} trains the general policy to control the 3-inch and 5-inch quadcopters. Uniform distributions were designed to encompass parameters from both sizes.
\begin{table}
\centering
\scriptsize
\caption{Randomization scheme for the general policy}
\begin{tabular}{|l|l|l|l|}
\hline
\textbf{Parameter} & \textbf{Distribution} & \textbf{Parameter} & \textbf{Distribution} \\ \hline
$\omega_{\text{min}}$ & $\mathcal{U}(0, 500)$ & $\hat{k}_{\omega}$ & $\mathcal{U}(10, 30)$ \\ \hline
$\omega_{\text{max}}$ & $\mathcal{U}(3000, 5000)$ & $\hat{k}_{x}$ & $\mathcal{U}(0.1, 0.3)$ \\ \hline
$k_l$ & $\mathcal{U}(0, 1)$ & $\hat{k}_{y}$ & $\mathcal{U}(0.1, 0.3)$ \\ \hline
$\tau$ & $\mathcal{U}(0.01, 0.1)$ & & \\ 
\hline
$\hat{k}_{p}$ & $\mathcal{U}(200, 800)$ & $\hat{k}_{p1}, \hat{k}_{p2}, \hat{k}_{p3}, \hat{k}_{p4}$ & $\hat{k}_{p} \pm \mathcal{U}(50)$ \\ \hline
$\hat{k}_{q}$ & $\mathcal{U}(200, 800)$ & $\hat{k}_{q1}, \hat{k}_{q2}, \hat{k}_{q3}, \hat{k}_{q4}$ & $\hat{k}_{q} \pm \mathcal{U}(50)$ \\ \hline
$\hat{k}_{r}$ & $\mathcal{U}(20, 80)$ & $\hat{k}_{r1}, \hat{k}_{r2}, \hat{k}_{r3}, \hat{k}_{r4}$ & $\hat{k}_{r}$ \\ \hline
$\hat{k}_{rd}$ & $\mathcal{U}(2, 8)$ & $\hat{k}_{r5}, \hat{k}_{r6}, \hat{k}_{r7}, \hat{k}_{r8}$ & $\hat{k}_{rd}$ \\ \hline
\end{tabular}
\label{tab:randomization_parameters}
\end{table}
\subsubsection{Fine tuned policy} \label{sec:finetuned}
We train five policies for the 3- and 5 inch quadcopter by applying randomization to the values from Tab. \ref{tab:param}, introducing variations of 0\%, 10\%, 20\%, and 30\%. Specifically, each parameter is multiplied by $\mathcal{U}(1 - p, 1 + p)$, where $p$ represents the randomization percentage. The thrust linearization constant $k_l$ is an exception, as it is capped to ensure it remains below 1.
\subsection{Policy}
The selected neural policy is a three-layer fully connected network with ReLU activation functions, with 64 neurons per layer (see Sec. \ref{sec:selecting_arc} for motivation). The policy takes in 20 observations, including the quadcopter's state and information about current and future gates, similar to prior work \cite{robinRL}:
\begin{align} \label{eq:observation}
    \mb{x}_{obs} = [\mb{p}^{g_i}
, \mb{v}^{g_i}, \boldsymbol \lambda^{g_i}, \mb{\Omega}, \mb{\boldsymbol \omega}, \mb{p}_{g_{i+1}}^{g_{i}}, \psi_{g_{i+1}}^{g_i}]^T
\end{align}
Here, ${g_i}$ denotes the reference frame of the $i-$th gate, and $\mb{p}_{g{i+1}}$ and $\psi_{g_{i+1}}$ represent the position and orientation of the next gate. The network outputs four motor commands ($\mb{u}$). Because domain randomization is applied to the model parameters, these parameters become part of the state in the Markov decision process (MDP). As a result, the full state is not completely observable with the current observation formulation \cite{tiboni2023domain}. Some approaches try to address this by incorporating state or action history into the observation \cite{Kaufmann2023, geles2024demonstrating}, while others use parameter inputs \cite{zhang_learning_2023}. In Sec. \ref{sec:selecting_arc}, we explore these methods and justify our choice of Eq. \ref{eq:observation}.

\section{EXPERIMENTAL SETUP}\label{section:experiment}
We test our policies on two quadcopters. Common between them is the compute hardware and experimental infrastructure: we use an STM32H743 microcontroller running INDIflight\footnote{\url{https://github.com/tudelft/indiflight}}, a fork of Betaflight.
An onboard EKF fuses data from a TDK InvenSense ICM-42688-P IMU, and position/attitude observations from an external optical motion capture system (Optitrack). The state is then used by the Neural Network, onboard, at an update rate of $1000$Hz. 
\begin{figure}[h]
    \centering
    \includegraphics[width=\linewidth]{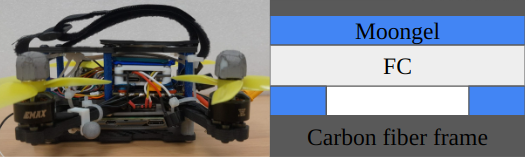}
    \caption{This picture shows the drone used for testing. Instead of bolts, four cornerpieces of Moongel are placed on top of the base frame, on which a 3D printed casing with the flight controller is placed. A second layer of Moongel is then placed on the casing, which is then slightly compressed downward by means of two zip-ties.}
    \label{fig:moongel}
\end{figure}
\\
During initial testing, at high speeds, propeller vibrations caused the accelerometer to saturate, exceeding the ±16g limit of the accelerometer. We aimed to address this problem by adding a vibration damping pad as shown in Fig. \ref{fig:moongel}. Sorbothane, adhesive tape and Moongel Damper Pads \cite{RTOM_Moongel} were compared, but the latter proved to provide the most effective high-frequency resonance control whilst still transmitting the lower frequency true movements of the drone. This result aligns with findings from previous research on damper pad performance, where Moongel was indeed shown to be optimal at eliminating noise \cite{GILANI2017103}. The 3D-printed casing, while not mandatory, aids in distributing the forces more evenly across the FC, whereas the zip-ties connected to the base frame provide preloaded compression for maintaining stability and improving damping performance.

\section{RESULTS}\label{section:results}
\subsection{Selecting Architecture} \label{sec:selecting_arc}
\begin{figure}
    \centering
    \includegraphics[width=\linewidth]{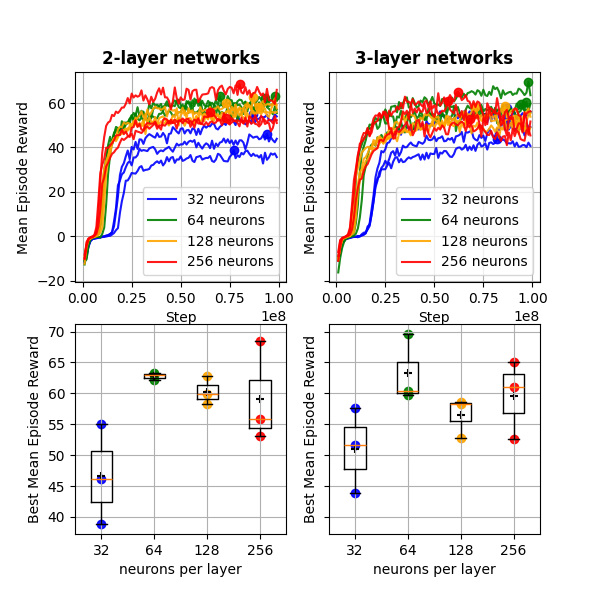}
    \caption{Comparison of network sizes: The 64,64,64 architecture achieved the highest mean episode reward among the evaluated architectures, based on three independent runs with 100 million time steps each.}
    \label{fig:size_comparison}
\end{figure}

\begin{figure}
    \centering
    \includegraphics[width=\linewidth]{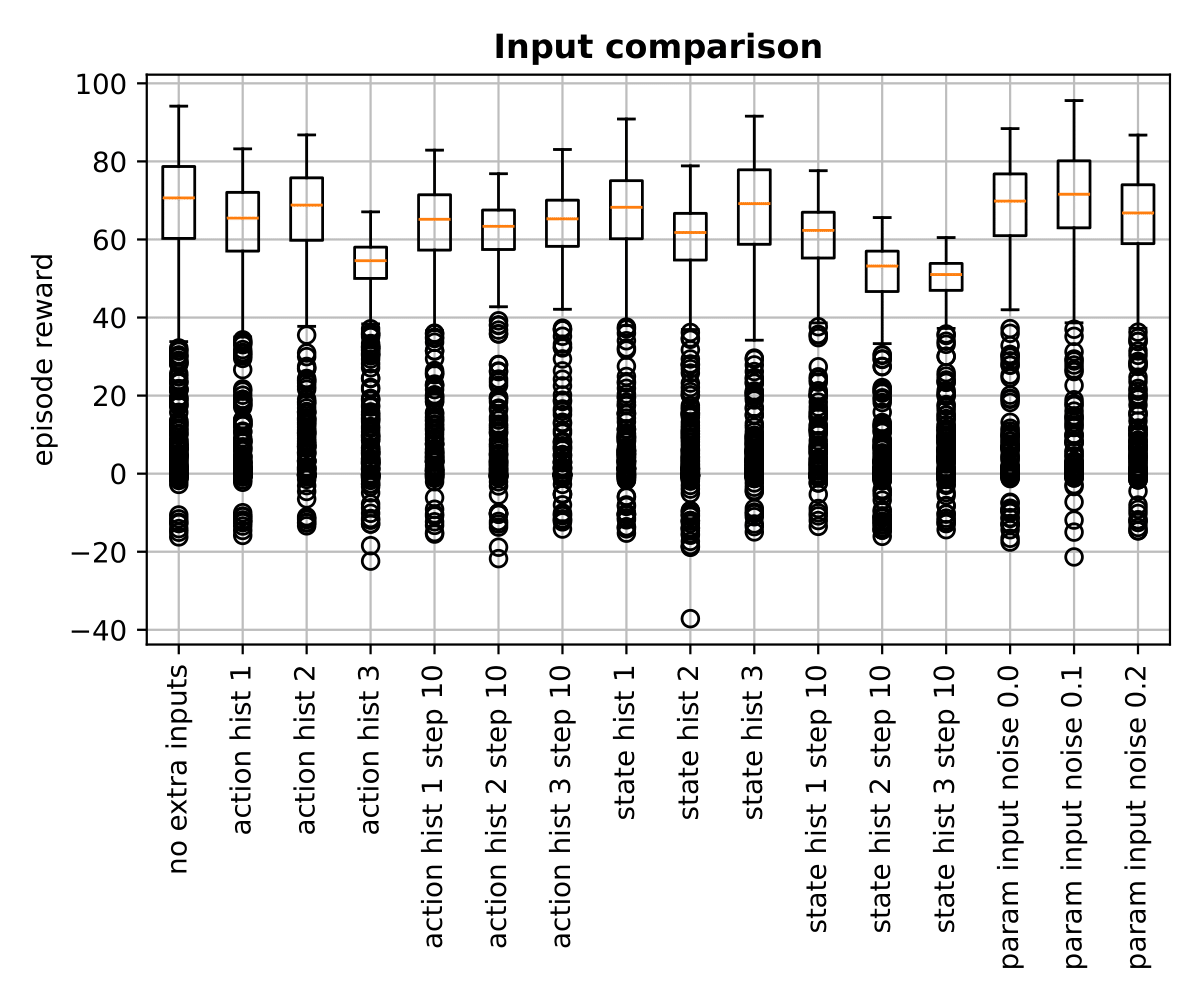}
    \caption{Comparison of training performance with various input modifications to the 64,64,64 ReLU network.}
    \label{fig:input_comparison}
\end{figure}

We evaluated several architectures using the randomized simulator and selected the 3-layer 64-neuron architecture as the best based on the highest obtained mean episode rewards from three independent runs, each consisting of 100 million time steps. Network size comparisons are illustrated in Fig. \ref{fig:size_comparison}.
Next, we modified the selected architecture to address partial observability by adding various inputs. The modifications included:
\begin{itemize}
    \item Adding the last 1, 2, or 3 actions to the observation
    \item Adding the last 1, 2, or 3 actions with stepsize 10 (steps of 0.1s in the past)
    \item Adding state history (last 1, 2, or 3 states, with or without stepsize 10)
    \item Adding model parameters (ground truth) or with 10\% and 20\% noise
\end{itemize}
Each variation followed the same training setup and was evaluated over 1000 rollouts. See Fig. \ref{fig:input_comparison} for results.
None of the modified inputs showed a clear improvement, leading us to exclude them from further use. This unexpected result indicates a need for further investigation, which may involve longer training periods, hyperparameter tuning, or alternative parameter representations. For the remainder of this paper, we will focus on the network without additional inputs.

\subsection{Simulation}
We trained the selected architecture on 9 different environments: the general randomization described in Sec. \ref{sec:general} and the fine-tuned models described in Sec. \ref{sec:finetuned}, which include 3-inch and 5-inch drones with 0\%, 10\%, 20\%, and 30\% randomization levels. Training followed the same procedure as above. Performance was evaluated with 1000 rollouts on both the 3-inch and 5-inch drone environments (with identical initial conditions sampled from the ranges described in \ref{sec:RLproblem}). See Tab. \ref{tab:sim_study}.
\begin{table}
\centering
\scriptsize
\caption{Simulation results averaged over 1000 policy rollouts}
\begin{tabular}{|c|c|c|c|c|c|c|c|c|c|}
\hline
\textbf{network} & \multicolumn{4}{|c|}{\textbf{3inch sim}} & \multicolumn{4}{|c|}{\textbf{5inch sim}} \\
\hline
 & \rotatebox{90}{ep rew} & \rotatebox{90}{ep len} & \rotatebox{90}{\#gates} & \rotatebox{90}{\%crash} & \rotatebox{90}{ep rew} & \rotatebox{90}{ep len} & \rotatebox{90}{\#gates} & \rotatebox{90}{\%crash} \\
\hline
\textbf{general} & \cellcolor{green!25}64.67 & \cellcolor{green!25}1140 & \cellcolor{green!25}40 & \cellcolor{green!25}5.4 & \cellcolor{green!25}76.78 & \cellcolor{green!25}1186 & \cellcolor{green!25}52 & \cellcolor{green!25}1.2 \\
\hline
\hline
\textbf{3inch 30\%} & \cellcolor{green!25}59.97 & \cellcolor{green!25}1200 & \cellcolor{green!25}41 & \cellcolor{green!25}0.0 & \cellcolor{red!25}3.95 & \cellcolor{red!25}204 & \cellcolor{red!25}4 & \cellcolor{red!25}99.8 \\
\hline
\textbf{3inch 20\%} & \cellcolor{green!25}69.02 & \cellcolor{green!25}1200 & \cellcolor{green!25}43 & \cellcolor{green!25}0.0 & \cellcolor{red!25}-6.24 & \cellcolor{red!25}127 & \cellcolor{red!25}3 & \cellcolor{red!25}100 \\
\hline
\textbf{3inch 10\%} & \cellcolor{green!25}77.59 & \cellcolor{green!25}1197 & \cellcolor{green!25}46 & \cellcolor{green!25}0.3 & \cellcolor{red!25}-2.24 & \cellcolor{red!25}70 & \cellcolor{red!25}2 & \cellcolor{red!25}100 \\
\hline
\textbf{3inch  0\%} & \cellcolor{green!25}78.55 & \cellcolor{green!25}1193 & \cellcolor{green!25}46 & \cellcolor{green!25}0.6 & \cellcolor{red!25}-1.78 & \cellcolor{red!25}89 & \cellcolor{red!25}2 & \cellcolor{red!25}100 \\
\hline
\hline
\textbf{5inch 30\%} & \cellcolor{red!25}4.23 & \cellcolor{red!25}195 & \cellcolor{red!25}3 & \cellcolor{red!25}100 & \cellcolor{green!25}74.03 & \cellcolor{green!25}1199 & \cellcolor{green!25}49 & \cellcolor{green!25}0.1 \\
\hline
\textbf{5inch 20\%} & \cellcolor{red!25}-0.35 & \cellcolor{red!25}125 & \cellcolor{red!25}1 & \cellcolor{red!25}100 & \cellcolor{green!25}88.33 & \cellcolor{green!25}1197 & \cellcolor{green!25}58 & \cellcolor{green!25}0.3 \\
\hline
\textbf{5inch 10\%} & \cellcolor{red!25}4.57 & \cellcolor{red!25}218 & \cellcolor{red!25}4 & \cellcolor{red!25}100 & \cellcolor{green!25}92.76 & \cellcolor{green!25}1183 & \cellcolor{green!25}55 & \cellcolor{green!25}1.5 \\
\hline
\textbf{5inch  0\%} & \cellcolor{red!25}-0.69 & \cellcolor{red!25}143 & \cellcolor{red!25}1 & \cellcolor{red!25}100 & \cellcolor{green!25}99.35 & \cellcolor{green!25}1179 & \cellcolor{green!25}59 & \cellcolor{green!25}2.0 \\
\hline
\end{tabular}
\label{tab:sim_study}
\end{table}

As shown, the general model performs effectively across both quadcopter simulations, navigating gates with a crash rate of 5.4\% in the 3-inch environment and 1.2\% in the 5-inch environment. However, it underperforms slightly compared to the fine-tuned models for each specific drone. Interestingly, the fine-tuned models do not transfer between drone types, resulting in 100\% crash rates due to non-overlapping parameter ranges, even with 30\% randomization. Among the fine-tuned models, lower randomization levels correlate with higher mean episode rewards, though at the cost of a slight increase in crash rates.
\begin{figure*}
    \centering
    \includegraphics[width=\textwidth]{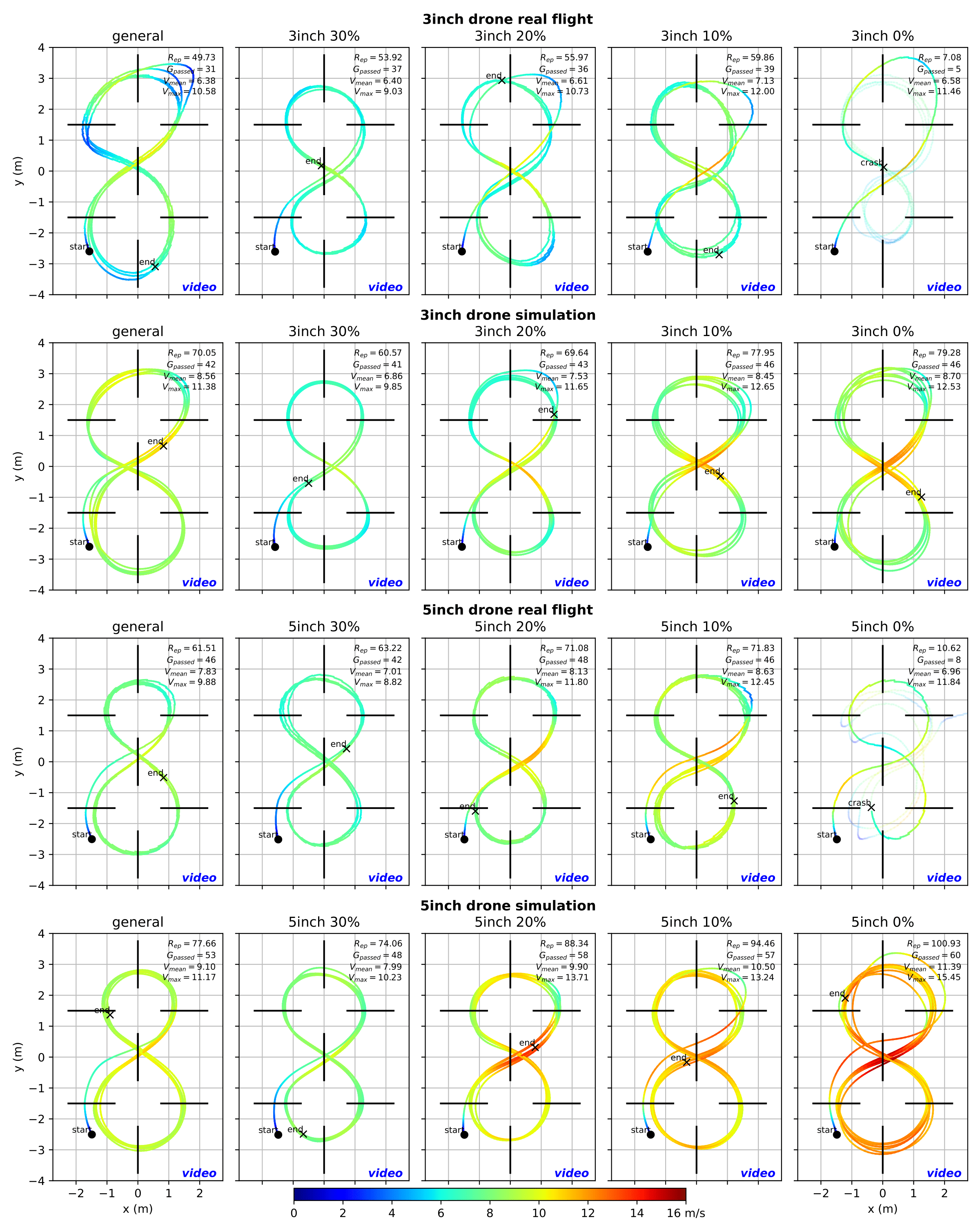}
    \caption{Comparison of real-world and simulated flight trajectories for 3-inch and 5-inch quadcopters, using the general policy (for both platforms) and fine-tuned policies (trained on their respective parametric models) with 30\%, 20\%, 10\%, and 0\% randomization}
    \label{fig:bigfig}
\end{figure*}


\begin{table*}
    \centering
    \scriptsize
    \begin{minipage}{0.48\textwidth}
    \centering
    \label{tab:pef_3inch}
    \caption{Performance measures: 3-inch flights}

\begin{tabular}{|c|c|c|c|c|c|c|c|c|}
\hline
\textbf{net}&\multicolumn{4}{|c|}{\textbf{Real Flight}} & \multicolumn{4}{|c|}{\textbf{Simulation}} \\
\hline
&\rotatebox{90}{ep rew} & \rotatebox{90}{\#gates} & \rotatebox{90}{$V_{\text{mean}}$} & \rotatebox{90}{$V_{\text{max}}$}
 & \rotatebox{90}{ep rew} & \rotatebox{90}{\#gates} & \rotatebox{90}{$V_{\text{mean}}$} & \rotatebox{90}{$V_{\text{max}}$} \\
\hline
\hline
\multirow{3}{*}{\rotatebox{90}{\textbf{Gen}}}
& \cellcolor{green!25}49.73 & \cellcolor{green!25}31& \cellcolor{green!25}6.38 & \cellcolor{green!25}10.58 & \cellcolor{green!25}69.76 & \cellcolor{green!25}42& \cellcolor{green!25}8.59 & \cellcolor{green!25}10.99 \\
& \cellcolor{green!25}49.38 & \cellcolor{green!25}31& \cellcolor{green!25}6.38 & \cellcolor{green!25}10.33 & \cellcolor{green!25}69.77 & \cellcolor{green!25}42& \cellcolor{green!25}8.59 & \cellcolor{green!25}10.96 \\
& \cellcolor{green!25}48.78 & \cellcolor{green!25}31& \cellcolor{green!25}6.31 & \cellcolor{green!25}10.4 & \cellcolor{green!25}69.97 & \cellcolor{green!25}42& \cellcolor{green!25}8.6 & \cellcolor{green!25}10.94 \\
\hline
\hline
\multirow{3}{*}{\rotatebox{90}{\textbf{30\%}}}
& \cellcolor{green!25}53.92 & \cellcolor{green!25}37& \cellcolor{green!25}6.4 & \cellcolor{green!25}9.03 & \cellcolor{green!25}60.44 & \cellcolor{green!25}41& \cellcolor{green!25}6.86 & \cellcolor{green!25}9.76 \\
& \cellcolor{green!25}53.48 & \cellcolor{green!25}37& \cellcolor{green!25}6.34 & \cellcolor{green!25}8.73 & \cellcolor{green!25}60.61 & \cellcolor{green!25}41& \cellcolor{green!25}6.86 & \cellcolor{green!25}9.68 \\
& \cellcolor{green!25}53.25 & \cellcolor{green!25}37& \cellcolor{green!25}6.33 & \cellcolor{green!25}8.85 & \cellcolor{green!25}60.6 & \cellcolor{green!25}41& \cellcolor{green!25}6.86 & \cellcolor{green!25}9.67 \\
\hline
\multirow{3}{*}{\rotatebox{90}{\textbf{20\%}}}
& \cellcolor{green!25}55.97 & \cellcolor{green!25}36& \cellcolor{green!25}6.61 & \cellcolor{green!25}10.73 & \cellcolor{green!25}69.67 & \cellcolor{green!25}43& \cellcolor{green!25}7.56 & \cellcolor{green!25}11.58 \\
& \cellcolor{green!25}55.4 & \cellcolor{green!25}35& \cellcolor{green!25}6.54 & \cellcolor{green!25}10.71 & \cellcolor{green!25}69.67 & \cellcolor{green!25}43& \cellcolor{green!25}7.55 & \cellcolor{green!25}11.58 \\
& \cellcolor{green!25}55.32 & \cellcolor{green!25}35& \cellcolor{green!25}6.49 & \cellcolor{green!25}10.48 & \cellcolor{green!25}69.72 & \cellcolor{green!25}43& \cellcolor{green!25}7.56 & \cellcolor{green!25}11.58 \\
\hline
\multirow{3}{*}{\rotatebox{90}{\textbf{10\%}}}
& \cellcolor{green!25}59.86 & \cellcolor{green!25}39& \cellcolor{green!25}7.13 & \cellcolor{green!25}12& \cellcolor{green!25}77.74 & \cellcolor{green!25}46& \cellcolor{green!25}8.45 & \cellcolor{green!25}13.15 \\
& \cellcolor{green!25}59.99 & \cellcolor{green!25}39& \cellcolor{green!25}7.13 & \cellcolor{green!25}11.56 & \cellcolor{green!25}78.17 & \cellcolor{green!25}46& \cellcolor{green!25}8.46 & \cellcolor{green!25}13.24 \\
& \cellcolor{green!25}59.34 & \cellcolor{green!25}38& \cellcolor{green!25}7.05 & \cellcolor{green!25}11.39 & \cellcolor{green!25}78.11 & \cellcolor{green!25}46& \cellcolor{green!25}8.46 & \cellcolor{green!25}13.25 \\
\hline
\multirow{3}{*}{\rotatebox{90}{\textbf{0\%}}}
& \cellcolor{red!25}7.08 & \cellcolor{red!25}5& \cellcolor{red!25}6.58 & \cellcolor{red!25}11.46 & \cellcolor{green!25}80.82 & \cellcolor{green!25}46& \cellcolor{green!25}8.79 & \cellcolor{green!25}12.8 \\
& \cellcolor{red!25}7.11 & \cellcolor{red!25}5& \cellcolor{red!25}6.49 & \cellcolor{red!25}11.25 & \cellcolor{green!25}81.09 & \cellcolor{green!25}46& \cellcolor{green!25}8.78 & \cellcolor{green!25}12.49 \\
& \cellcolor{red!25}10.18 & \cellcolor{red!25}7& \cellcolor{red!25}6.51 & \cellcolor{red!25}10.87 & \cellcolor{green!25}80.93 & \cellcolor{green!25}46& \cellcolor{green!25}8.77 & \cellcolor{green!25}12.38 \\
\hline\end{tabular}

    \end{minipage}
    \hfill
    \begin{minipage}{0.48\textwidth}
    \centering
    \label{tab:perf_5inch}
    \caption{Performance measures: 5-inch flights}
\begin{tabular}{|c|c|c|c|c|c|c|c|c|}
\hline
\textbf{net}&\multicolumn{4}{|c|}{\textbf{Real Flight}} & \multicolumn{4}{|c|}{\textbf{Simulation}} \\
\hline
&\rotatebox{90}{ep rew} & \rotatebox{90}{\#gates} & \rotatebox{90}{$V_{\text{mean}}$} & \rotatebox{90}{$V_{\text{max}}$}
 & \rotatebox{90}{ep rew} & \rotatebox{90}{\#gates} & \rotatebox{90}{$V_{\text{mean}}$} & \rotatebox{90}{$V_{\text{max}}$} \\
\hline
\hline
\multirow{3}{*}{\rotatebox{90}{\textbf{Gen}}}
& \cellcolor{green!25}61.51 & \cellcolor{green!25}46& \cellcolor{green!25}7.83 & \cellcolor{green!25}9.88 & \cellcolor{green!25}77.14 & \cellcolor{green!25}53& \cellcolor{green!25}9.09 & \cellcolor{green!25}11.03 \\
& \cellcolor{green!25}61.22 & \cellcolor{green!25}46& \cellcolor{green!25}7.84 & \cellcolor{green!25}9.83 & \cellcolor{green!25}77.5 & \cellcolor{green!25}53& \cellcolor{green!25}9.12 & \cellcolor{green!25}11.37 \\
& \cellcolor{green!25}60.65 & \cellcolor{green!25}46& \cellcolor{green!25}7.8 & \cellcolor{green!25}9.84 & \cellcolor{green!25}77.74 & \cellcolor{green!25}53& \cellcolor{green!25}9.14 & \cellcolor{green!25}11.26 \\
\hline
\hline
\multirow{3}{*}{\rotatebox{90}{\textbf{30\%}}}
& \cellcolor{green!25}63.22 & \cellcolor{green!25}42& \cellcolor{green!25}7.01 & \cellcolor{green!25}8.82 & \cellcolor{green!25}74.38 & \cellcolor{green!25}48& \cellcolor{green!25}8.01 & \cellcolor{green!25}10.22 \\
& \cellcolor{green!25}62.87 & \cellcolor{green!25}42& \cellcolor{green!25}7.01 & \cellcolor{green!25}8.73 & \cellcolor{green!25}74.1 & \cellcolor{green!25}48& \cellcolor{green!25}8.01 & \cellcolor{green!25}10.15 \\
& \cellcolor{green!25}62.59 & \cellcolor{green!25}41& \cellcolor{green!25}6.98 & \cellcolor{green!25}8.7 & \cellcolor{green!25}73.7 & \cellcolor{green!25}48& \cellcolor{green!25}8& \cellcolor{green!25}10.1 \\
\hline
\multirow{3}{*}{\rotatebox{90}{\textbf{20\%}}}
& \cellcolor{green!25}71.08 & \cellcolor{green!25}48& \cellcolor{green!25}8.13 & \cellcolor{green!25}11.8 & \cellcolor{green!25}89.44 & \cellcolor{green!25}58& \cellcolor{green!25}9.99 & \cellcolor{green!25}13.8 \\
& \cellcolor{green!25}71.44 & \cellcolor{green!25}49& \cellcolor{green!25}8.16 & \cellcolor{green!25}11.74 & \cellcolor{green!25}89.56 & \cellcolor{green!25}58& \cellcolor{green!25}9.99 & \cellcolor{green!25}13.9 \\
& \cellcolor{green!25}71.3 & \cellcolor{green!25}48& \cellcolor{green!25}8.13 & \cellcolor{green!25}11.52 & \cellcolor{green!25}89.84 & \cellcolor{green!25}58& \cellcolor{green!25}10& \cellcolor{green!25}14.09 \\
\hline
\multirow{3}{*}{\rotatebox{90}{\textbf{10\%}}}
& \cellcolor{green!25}71.83 & \cellcolor{green!25}46& \cellcolor{green!25}8.63 & \cellcolor{green!25}12.45 & \cellcolor{green!25}95.68 & \cellcolor{green!25}58& \cellcolor{green!25}10.53 & \cellcolor{green!25}12.81 \\
& \cellcolor{green!25}72.67 & \cellcolor{green!25}47& \cellcolor{green!25}8.73 & \cellcolor{green!25}12.18 & \cellcolor{green!25}94.93 & \cellcolor{green!25}58& \cellcolor{green!25}10.52 & \cellcolor{green!25}12.81 \\
& \cellcolor{green!25}71.98 & \cellcolor{green!25}46& \cellcolor{green!25}8.63 & \cellcolor{green!25}12.07 & \cellcolor{green!25}95.67 & \cellcolor{green!25}58& \cellcolor{green!25}10.54 & \cellcolor{green!25}13.05 \\
\hline
\multirow{3}{*}{\rotatebox{90}{\textbf{0\%}}}
& \cellcolor{red!25}10.62 & \cellcolor{red!25}8& \cellcolor{red!25}6.96 & \cellcolor{red!25}11.84 & \cellcolor{green!25}99.9 & \cellcolor{green!25}60& \cellcolor{green!25}11.34 & \cellcolor{green!25}15.4 \\
& \cellcolor{red!25}17.16 & \cellcolor{red!25}13& \cellcolor{red!25}7.46 & \cellcolor{red!25}13.13 & \cellcolor{green!25}100.55 & \cellcolor{green!25}59& \cellcolor{green!25}11.34 & \cellcolor{green!25}15.53 \\
& \cellcolor{red!25}11.34 & \cellcolor{red!25}8& \cellcolor{red!25}7.43 & \cellcolor{red!25}12.39 & \cellcolor{green!25}97.48 & \cellcolor{green!25}59& \cellcolor{green!25}11.27 & \cellcolor{green!25}15.37 \\
\hline\end{tabular}
    \end{minipage}
\end{table*}
\subsection{Real flights}
\begin{figure}
    \centering
    \includegraphics[width=\linewidth]{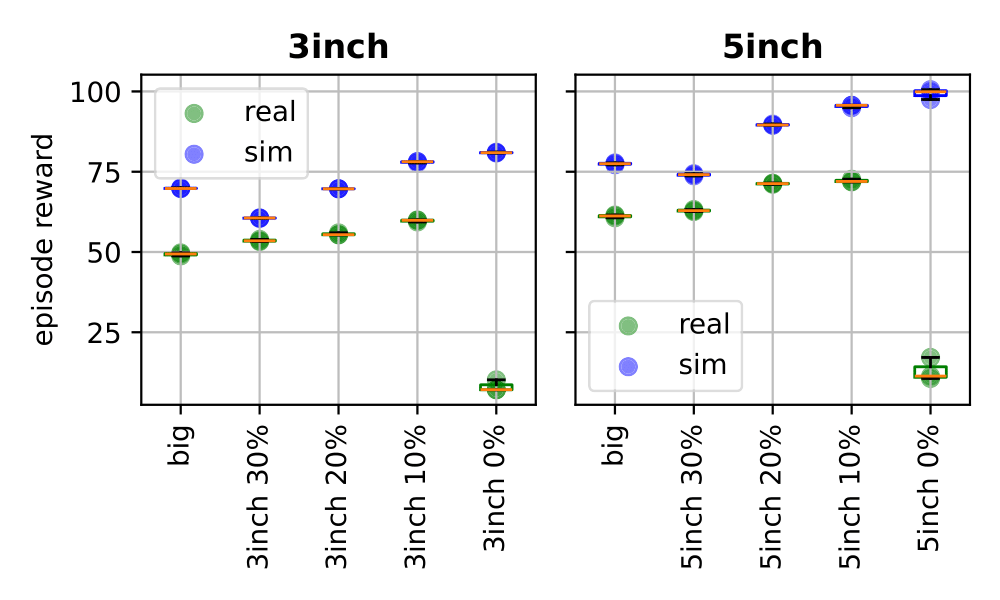}
    \caption{Domain randomization enables the same network to fly different quadcopters; however with performance penalties. No domain randomization results in failed sim2real transfer: our DR approach stands as an alternative to higher modeling effort, at a small performance cost.}
    \label{fig:rew_boxplot}
\end{figure}

We now test the networks that performed well in simulation on their respective platforms. Specifically, fine-tuned models are evaluated on their designated platforms, while the general model is tested on both the 3-inch and 5-inch quadcopters.  All flights begin from a hover state, 1 meter in front of gate 1, and are repeated three times (12 seconds each, matching the maximum episode length used in the RL framework). The first flight uses a fully charged battery, with subsequent flights performed as the battery depletes, leaving it near empty by the third run. Flight logs are subsampled at 100Hz to match the simulator's time step, enabling direct comparison through the episode reward metric. Additionally, each flight is replicated in the simulator using the same initial conditions. The trajectories of both real and simulated flights are shown in Fig. \ref{fig:bigfig}. It can be seen here that the general network successfully navigates the race track on both drones, albeit slightly slower. Performance metrics for all flights are presented in Tab. \ref{tab:pef_3inch} and \ref{tab:perf_5inch}, showing that as randomization decreases from 30\% to 10\%, performance improves, with the highest rewards at 10\%. At 0\% randomization, the drone no longer passes through the gates. Fig.~\ref{fig:rew_boxplot} provides a box plot of the episode rewards, clearly illustrating how the reality gap widens as domain randomization decreases.

\subsection{Time Optimality}
The rewards in the framework are designed to learn time-optimal flight. It is useful to not only compare the achieved episode rewards to the optimal during training but also compare actual lap times to time-optimal flight as a means of validation and context.
\begin{figure}
    \centering
    \includegraphics[width=0.5\linewidth]{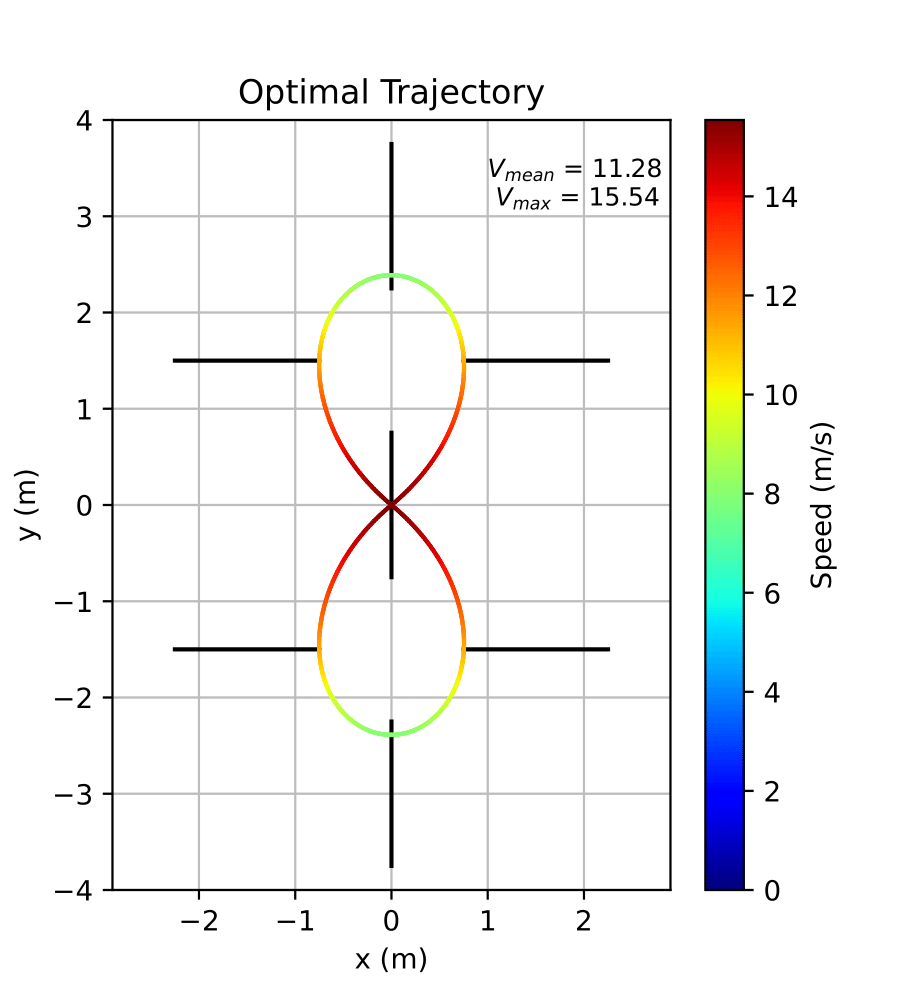}
    \caption{The optimal control solution through the race track for the 5inch quadcopter}
    \label{fig:optimal_control}
    \vspace{-4mm} 
\end{figure}
The optimal figure 8 path (see Fig.\ref{fig:optimal_control}) was formulated as a boundary value problem and solved with \cite{Caillau_OptimalControl_jl_a_Julia} for the 5-inch model, yielding a theoretical lap-time of $1.07$s, compared to $1.6$s for our fastest RL policies in simulation, and $1.9$s in real-life.
Even though the time-optimal program did not include drag terms, we attribute the bulk of the difference to the sub-optimal path taken by the RL policies; relatively far away from the edges of the gates. However, the optimal control approach returns $1.53$s when forcing passing the gate-centers. This suggests that our reward function does not perfectly align with optimizing lap times. As shown in Tab. \ref{tab:perf_5inch}, the 10\% fine-tuned network achieves the highest real-life performance in terms of episode reward, yet it flies through fewer gates than the 20\% variant. We therefore conclude that, to achieve more time-optimal performance on a tight track, the reward function should include terms that allow better tuning of the risk-reward balance, encouraging the agent to fly closer to the gate edges.\\
\section{CONCLUSION} \label{sec:conclusion}
In this work, we introduced a neural network controller for high-speed drone racing capable of operating across multiple quadcopter platforms. We demonstrated its effectiveness with real flights on both 3-inch and 5-inch quadcopters at speeds up to 10 m/s.
Our benchmarking involved comparing this generalized controller against neural policies specifically trained for the 3-inch and 5-inch quadcopters with varying levels of domain randomization. While our generalized model showed slightly reduced performance compared to the fine-tuned controllers, it exhibited superior transferability across platforms.
Our experiments revealed that increased domain randomization initially enhances robustness and helps bridge the reality gap. However, excessive randomization ultimately compromises flight speed.
We explored incorporating online adaptation into our network, as discussed in \cite{blaha_control_2024} and \cite{zhang_learning_2023}, but did not achieve the desired improvements. Potential factors include training duration, hyperparameter tuning, or the choice of parameter representations. Nevertheless, the significant success of our generalized approach, even without additional inputs, highlights its promise and paves the way for future research. Addressing the identified challenges could lead to substantial performance enhancements, making this an exciting avenue for further exploration.

\addtolength{\textheight}{-8cm}   










\bibliographystyle{IEEEtran}
\bibliography{root}

\begin{thebibliography}{10}
\providecommand{\url}[1]{#1}
\csname url@samestyle\endcsname
\providecommand{\newblock}{\relax}
\providecommand{\bibinfo}[2]{#2}
\providecommand{\BIBentrySTDinterwordspacing}{\spaceskip=0pt\relax}
\providecommand{\BIBentryALTinterwordstretchfactor}{4}
\providecommand{\BIBentryALTinterwordspacing}{\spaceskip=\fontdimen2\font plus
\BIBentryALTinterwordstretchfactor\fontdimen3\font minus \fontdimen4\font\relax}
\providecommand{\BIBforeignlanguage}[2]{{%
\expandafter\ifx\csname l@#1\endcsname\relax
\typeout{** WARNING: IEEEtran.bst: No hyphenation pattern has been}%
\typeout{** loaded for the language `#1'. Using the pattern for}%
\typeout{** the default language instead.}%
\else
\language=\csname l@#1\endcsname
\fi
#2}}
\providecommand{\BIBdecl}{\relax}
\BIBdecl

\bibitem{Hassanalian2017ClassificationsAA}
M.~Hassanalian and A.~Abdelkefi, ``Classifications, applications, and design challenges of drones: A review,'' \emph{Progress in Aerospace Sciences}, vol.~91, pp. 99--131, 2017.

\bibitem{hanover_autonomous_2024}
D.~Hanover, A.~Loquercio, L.~Bauersfeld, A.~Romero, R.~Penicka, Y.~Song, G.~Cioffi, E.~Kaufmann, and D.~Scaramuzza, ``Autonomous {{Drone Racing}}: {{A Survey}},'' \emph{IEEE Transactions on Robotics}, vol.~40, pp. 3044--3067, Jan. 2024.

\bibitem{Autonomous_Drone_Racing_with_Deep_Reinforcement_Learning}
Y.~Song, M.~Steinweg, E.~Kaufmann, and D.~Scaramuzza, ``Autonomous drone racing with deep reinforcement learning,'' in \emph{2021 IEEE/RSJ International Conference on Intelligent Robots and Systems (IROS)}.\hskip 1em plus 0.5em minus 0.4em\relax IEEE, 2021, pp. 1205--1212.

\bibitem{penicka2022learning}
R.~Penicka, Y.~Song, E.~Kaufmann, and D.~Scaramuzza, ``Learning minimum-time flight in cluttered environments,'' \emph{IEEE Robotics and Automation Letters}, vol.~7, no.~3, pp. 7209--7216, 2022.

\bibitem{OCvsRL}
\BIBentryALTinterwordspacing
Y.~Song, A.~Romero, M.~Müller, V.~Koltun, and D.~Scaramuzza, ``Reaching the limit in autonomous racing: Optimal control versus reinforcement learning,'' \emph{Science Robotics}, vol.~8, no.~82, p. eadg1462, 2023. [Online]. Available: \url{https://www.science.org/doi/abs/10.1126/scirobotics.adg1462}
\BIBentrySTDinterwordspacing

\bibitem{Kaufmann2023}
\BIBentryALTinterwordspacing
E.~Kaufmann, L.~Bauersfeld, A.~Loquercio, M.~M{\"u}ller, V.~Koltun, and D.~Scaramuzza, ``Champion-level drone racing using deep reinforcement learning,'' \emph{Nature}, vol. 620, no. 7976, pp. 982--987, Aug 2023. [Online]. Available: \url{https://doi.org/10.1038/s41586-023-06419-4}
\BIBentrySTDinterwordspacing

\bibitem{eschmann_learning_2023}
J.~Eschmann, D.~Albani, and G.~Loianno, ``Learning to {Fly} in {Seconds},'' Nov. 2023, arXiv:2311.13081.

\bibitem{josifovski2022analysis}
J.~Josifovski, M.~Malmir, N.~Klarmann, B.~L. {\v{Z}}agar, N.~Navarro-Guerrero, and A.~Knoll, ``Analysis of randomization effects on sim2real transfer in reinforcement learning for robotic manipulation tasks,'' in \emph{2022 IEEE/RSJ International Conference on Intelligent Robots and Systems (IROS)}.\hskip 1em plus 0.5em minus 0.4em\relax IEEE, 2022, pp. 10\,193--10\,200.

\bibitem{andrychowicz2020learning}
O.~M. Andrychowicz, B.~Baker, M.~Chociej, R.~Jozefowicz, B.~McGrew, J.~Pachocki, A.~Petron, M.~Plappert, G.~Powell, A.~Ray \emph{et~al.}, ``Learning dexterous in-hand manipulation,'' \emph{The International Journal of Robotics Research}, vol.~39, no.~1, pp. 3--20, 2020.

\bibitem{loquercio2019deep}
A.~Loquercio, E.~Kaufmann, R.~Ranftl, A.~Dosovitskiy, V.~Koltun, and D.~Scaramuzza, ``Deep drone racing: From simulation to reality with domain randomization,'' \emph{IEEE Transactions on Robotics}, vol.~36, no.~1, pp. 1--14, 2019.

\bibitem{tobin2017domain}
J.~Tobin, R.~Fong, A.~Ray, J.~Schneider, W.~Zaremba, and P.~Abbeel, ``Domain randomization for transferring deep neural networks from simulation to the real world,'' in \emph{2017 IEEE/RSJ international conference on intelligent robots and systems (IROS)}.\hskip 1em plus 0.5em minus 0.4em\relax IEEE, 2017, pp. 23--30.

\bibitem{robinSl}
R.~Ferede, G.~de~Croon, C.~De~Wagter, and D.~Izzo, ``End-to-end neural network based optimal quadcopter control,'' \emph{Robotics and Autonomous Systems}, vol. 172, p. 104588, 2024.

\bibitem{Seb}
S.~Origer, C.~De~Wagter, R.~Ferede, G.~C. de~Croon, and D.~Izzo, ``Guidance \& control networks for time-optimal quadcopter flight,'' \emph{arXiv preprint arXiv:2305.02705}, 2023.

\bibitem{robinRL}
R.~Ferede, C.~De~Wagter, D.~Izzo, and G.~C. De~Croon, ``End-to-end reinforcement learning for time-optimal quadcopter flight,'' in \emph{2024 IEEE International Conference on Robotics and Automation (ICRA)}.\hskip 1em plus 0.5em minus 0.4em\relax IEEE, 2024, pp. 6172--6177.

\bibitem{tiboni2023domain}
G.~Tiboni, P.~Klink, J.~Peters, T.~Tommasi, C.~D'Eramo, and G.~Chalvatzaki, ``Domain randomization via entropy maximization,'' \emph{arXiv preprint arXiv:2311.01885}, 2023.

\bibitem{molchanov2019sim}
A.~Molchanov, T.~Chen, W.~H{\"o}nig, J.~A. Preiss, N.~Ayanian, and G.~S. Sukhatme, ``Sim-to-(multi)-real: Transfer of low-level robust control policies to multiple quadrotors,'' in \emph{2019 IEEE/RSJ International Conference on Intelligent Robots and Systems (IROS)}.\hskip 1em plus 0.5em minus 0.4em\relax IEEE, 2019, pp. 59--66.

\bibitem{blaha_control_2024}
T.~Blaha, E.~Smeur, and B.~Remes, ``Control of {Unknown} {Quadrotors} from a {Single} {Throw},'' \emph{arXiv}, vol. arXiv:2406.11723, 2024.

\bibitem{smeur_adaptive_2016}
E.~J.~J. Smeur, Q.~Chu, and G.~C. H.~E. de~Croon, ``Adaptive {Incremental} {Nonlinear} {Dynamic} {Inversion} for {Attitude} {Control} of {Micro} {Air} {Vehicles},'' \emph{Journal of Guidance, Control, and Dynamics}, vol.~39, no.~3, pp. 450--461, Mar. 2016.

\bibitem{zhang_learning_2023}
D.~Zhang, A.~Loquercio, X.~Wu, A.~Kumar, J.~Malik, and M.~W. Mueller, ``Learning a {{Single Near-hover Position Controller}} for {{Vastly Different Quadcopters}},'' May 2023.

\bibitem{ppo}
J.~Schulman, F.~Wolski, P.~Dhariwal, A.~Radford, and O.~Klimov, ``Proximal policy optimization algorithms,'' \emph{arXiv preprint arXiv:1707.06347}, 2017.

\bibitem{stable-baselines3}
\BIBentryALTinterwordspacing
A.~Raffin, A.~Hill, A.~Gleave, A.~Kanervisto, M.~Ernestus, and N.~Dormann, ``Stable-baselines3: Reliable reinforcement learning implementations,'' \emph{Journal of Machine Learning Research}, vol.~22, no. 268, pp. 1--8, 2021. [Online]. Available: \url{http://jmlr.org/papers/v22/20-1364.html}
\BIBentrySTDinterwordspacing

\bibitem{geles2024demonstrating}
I.~Geles, L.~Bauersfeld, A.~Romero, J.~Xing, and D.~Scaramuzza, ``Demonstrating agile flight from pixels without state estimation,'' \emph{arXiv preprint arXiv:2406.12505}, 2024.

\bibitem{RTOM_Moongel}
{RTOM Corporation}, ``Moongel damper pads,'' \url{https://rtom.com/moongel-damper-pad/}, accessed: August 20, 2024.

\bibitem{GILANI2017103}
\BIBentryALTinterwordspacing
S.~F. ul~Haq~Gilani, M.~H. bin Mohd~Khir, R.~Ibrahim, E.~ul~Hassan~Kirmani, and S.~I. ul~Haq~Gilani, ``Modelling and development of a vibration-based electromagnetic energy harvester for industrial centrifugal pump application,'' \emph{Microelectronics Journal}, vol.~66, pp. 103--111, 2017. [Online]. Available: \url{https://www.sciencedirect.com/science/article/pii/S0026269216305948}
\BIBentrySTDinterwordspacing

\bibitem{Caillau_OptimalControl_jl_a_Julia}
\BIBentryALTinterwordspacing
J.-B. Caillau, O.~Cots, J.~Gergaud, P.~Martinon, and S.~Sed, ``{OptimalControl.jl: a Julia package to model and solve optimal control problems with ODE's}.'' [Online]. Available: \url{https://control-toolbox.org/OptimalControl.jl}
\BIBentrySTDinterwordspacing

\end{thebibliography}

\end{document}